\documentclass[10pt,twocolumn,letterpaper]{article}

\usepackage{iccv}              

\usepackage{cuted}       
\usepackage{caption}     

\usepackage{multirow}
\usepackage[normalem]{ulem}
\useunder{\uline}{\ul}{}

\usepackage{kotex} 
\usepackage{amsthm}
\usepackage{adjustbox}

\usepackage{./rpm_packages/rpm_SIunits}
\usepackage{./rpm_packages/rpm_acronyms}
\usepackage{./rpm_packages/rpm_math}
\usepackage{./rpm_packages/rpm_misc}
\usepackage[accsupp]{axessibility}  

\definecolor{iccvblue}{rgb}{0.21,0.49,0.74}
\usepackage[pagebackref,breaklinks,colorlinks,allcolors=iccvblue]{hyperref}

\def\paperID{5028} 
\def\confName{ICCV}
\def\confYear{2025}

\title{TRAN-D: 2D Gaussian Splatting-based Sparse-view Transparent Object \\ Depth Reconstruction via Physics Simulation for Scene Update}

\author{Jeongyun Kim$^*$\\
Seoul National University\\
{\tt\small jeongyunkim@snu.ac.kr}
\and
Seunghoon Jeong$^*$\\
Seoul National University\\
{\tt\small shoon0602@snu.ac.kr}
\and
Giseop Kim\\
DGIST\\
{\tt\small gsk@dgist.ac.kr}
\and
Myung-Hwan Jeon\\
Kumoh National Institute of Technology\\
{\tt\small mhjeon@kumoh.ac.kr}
\and
Eunji Jun\\
Hyundai Motor Group\\
{\tt\small ejjun@hyundai.com}
\and
Ayoung Kim\\
Seoul National University\\
{\tt\small ayoungk@snu.ac.kr}
}

\begin{document}
\maketitle
\begin{abstract}
Understanding the 3D geometry of transparent objects from RGB images is challenging due to their inherent physical properties, such as reflection and refraction. To address these difficulties, especially in scenarios with sparse views and dynamic environments, we introduce TRAN-D, a novel 2D Gaussian Splatting-based depth reconstruction method for transparent objects.
Our key insight lies in separating transparent objects from the background, enabling focused optimization of Gaussians corresponding to the object. We mitigate artifacts with an object‐aware loss that places Gaussians in obscured regions, ensuring coverage of invisible surfaces while reducing overfitting. Furthermore, we incorporate a physics-based simulation that refines the reconstruction in just a few seconds, effectively handling object removal and chain‐reaction movement of remaining objects without the need for rescanning.
TRAN-D is evaluated on both synthetic and real-world sequences, and it consistently demonstrated robust improvements over existing GS-based state-of-the-art methods. In comparison with baselines, TRAN-D reduces the mean absolute error by over 39\% for the synthetic TRansPose sequences. Furthermore, despite being updated using only one image, TRAN-D reaches a $\delta < 2.5$ cm accuracy of 48.46\%, over 1.5 times that of baselines, which uses six images. Code and more results are available at \href{https://jeongyun0609.github.io/TRAN-D/}{https://jeongyun0609.github.io/TRAN-D/}.
\vspace{-4mm}
\end{abstract}    
\def\thefootnote{*}\footnotetext{These authors contributed equally to this work}\def\thefootnote{\arabic{footnote}}
\section{Introduction}
\label{sec:intro}

\begin{figure}[!t]
    \centering
    \includegraphics[width=0.9\linewidth]{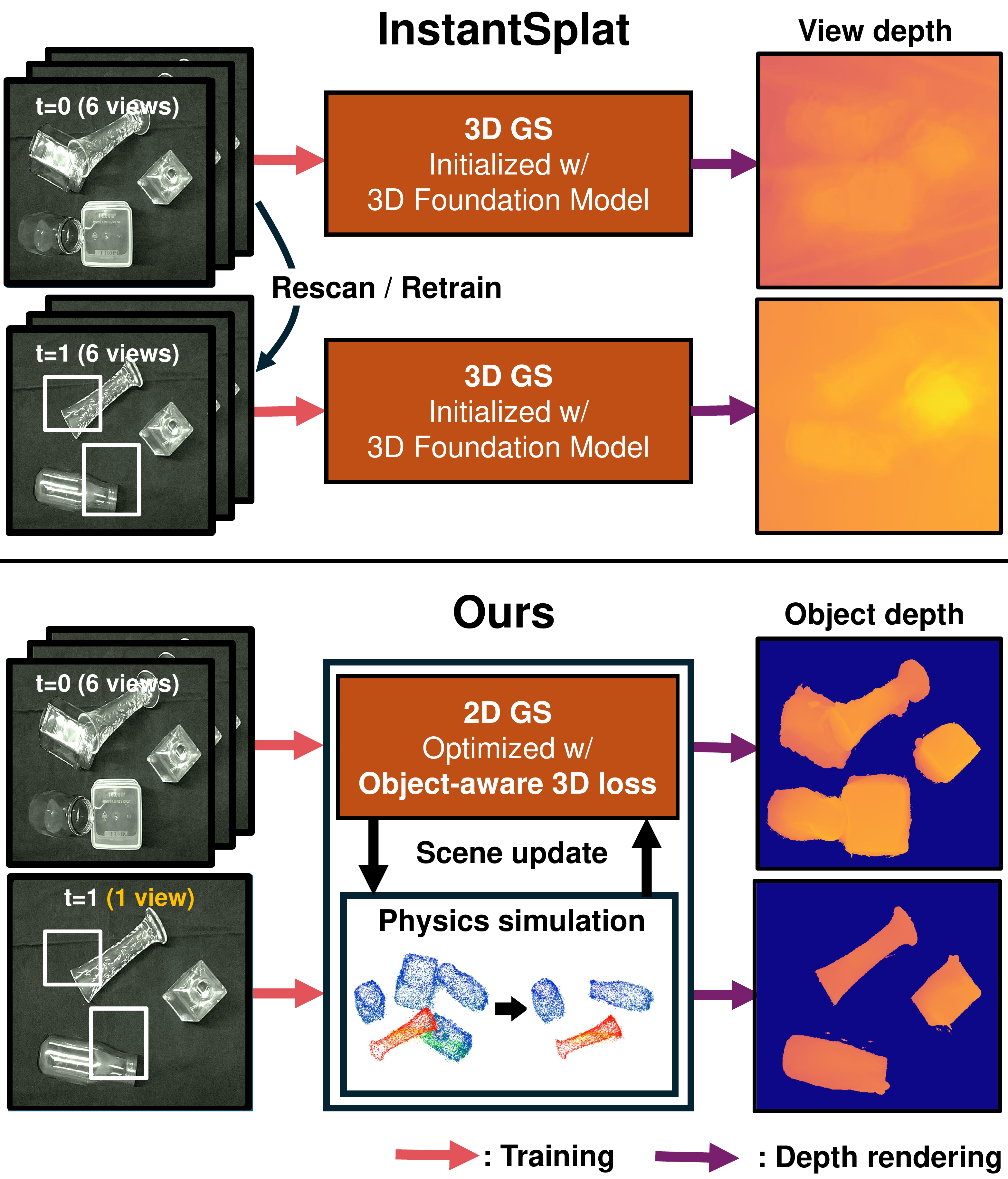}
    \vspace{-3mm}
    \caption{TRAN-D optimizes 2D Gaussians with object-aware 3D loss in sparse-view settings and refines their placement through physics simulation. Compared to baselines such as InstantSplat \cite{fan2024instantsplat}, our approach achieves more accurate depth reconstruction.}
    \label{fig:main}
    \vspace{-6mm}
\end{figure}

Transparent objects present unique challenges in computer vision due to their complex transmission, reflection, and refraction properties. 
Due to this difficulty, the 3D geometry of transparent objects has been underexplored, while most existing works on transparent objects handle 2D problems of segmentation \cite{xie-2020-eccv, mei-2022-cvpr} and detection \cite{jiang-2022-ral, mei-2020-cvpr}. 
In particular, reliable depth reconstruction for transparent objects remains an ill-posed problem, posing challenges for both conventional Time-of-Flight (ToF) sensors and recent neural rendering methods.
With the recent advent of volumetric neural rendering techniques like \ac{NeRF} \cite{mildenhall2021nerf} and \ac{GS} \cite{kerbl20233d}, researchers have started exploring 3D dense depth reconstruction for transparent objects.

To address the depth reconstruction problem for transparent objects, methods leveraging \ac{NeRF} \cite{ichnowski-2022-corl, kerr-2022-corl, duisterhof2024residualnerf, ummadisingu2024said} and \ac{GS} \cite{jykim-2025-icra} have been proposed. However, these methods require extensive training times and dense view inputs. Furthermore, they struggle with object dynamics; when objects move, the entire scene must be rescanned, making dense depth reconstruction highly time-consuming.

Recent advances in sparse-view \ac{NVS} \cite{huang2025fatesgs, tang2024hisplat, zhang2024transplat, fan2024instantsplat} have significantly reduced training times and alleviated the need for dense views by leveraging 3D foundation models \cite{wang2024dust3r, mast3r_eccv24} or depth estimation models \cite{Ranftl2022}. However, these methods still face challenges when applied to transparent objects. Due to generalization bias in foundation models, they often misinterpret the boundaries between transparent objects and backgrounds, which leads to inaccuracies in depth reconstruction.

In this work, we propose \textbf{TRAN-D}, a physics simulation-aided sparse-view \ac{2DGS} method for \textbf{\uline{TRAN}}sparent object \textbf{\uline{D}}epth reconstruction. Unlike existing approaches that struggle with view sparsity and object dynamics, TRAN-D builds upon a 2D Gaussian framework that effectively captures objects' geometric characteristics, ensuring accurate depth reconstruction, as shown in \figref{fig:main}.

A key component of TRAN-D is the use of segmentation-mask obtained through a Grounded SAM \cite{ren2024grounded} fine-tuned for transparent objects. By jointly splatting these features along with RGB values, TRAN-D focuses optimization on object regions while suppressing background interference, leading to more robust and precise depth reconstruction. Additionally, we introduce an object-aware 3D loss that optimizes Gaussian placement even in obscured regions, reducing overfitting and improving reconstruction quality. Furthermore, when objects are removed, a physics-based simulation updates the scene representation by relocating object-specific 2D Gaussians and refining the reconstruction from a single post-change image. This process enables seamless object removal and precise adaptation of the remaining scene, addressing the challenges posed by transparent object dynamics.

\begin{itemize}
\item  \textbf{Segmentation-Based Transparent Object Splatting: }
2D Gaussian optimization is enhanced by isolating transparent objects with segmentation masks, reducing background interference, and improving depth reconstruction accuracy. This focus on object-aware splatting boosts precision and streamlines the overall reconstruction process.

\item  \textbf{Object-Aware 3D Loss for Obscured Coverage: }
Object-aware 3D loss that strategically positions Gaussians in obscured regions is introduced. By ensuring a more uniform surface representation, this loss reduces overfitting, curtails the number of Gaussians required, and maintains reconstruction quality.

\item  \textbf{Physics Simulation for Object Dynamics: }
Physics-based simulation is incorporated to handle interaction occurring from object dynamics efficiently. By predicting object movements, we seamlessly adjust the 2D Gaussian representation, using minimal computational resources while preserving depth accuracy.

\end{itemize}

\begin{figure*}[ht]
    \centering
    \includegraphics[width=0.85\linewidth]{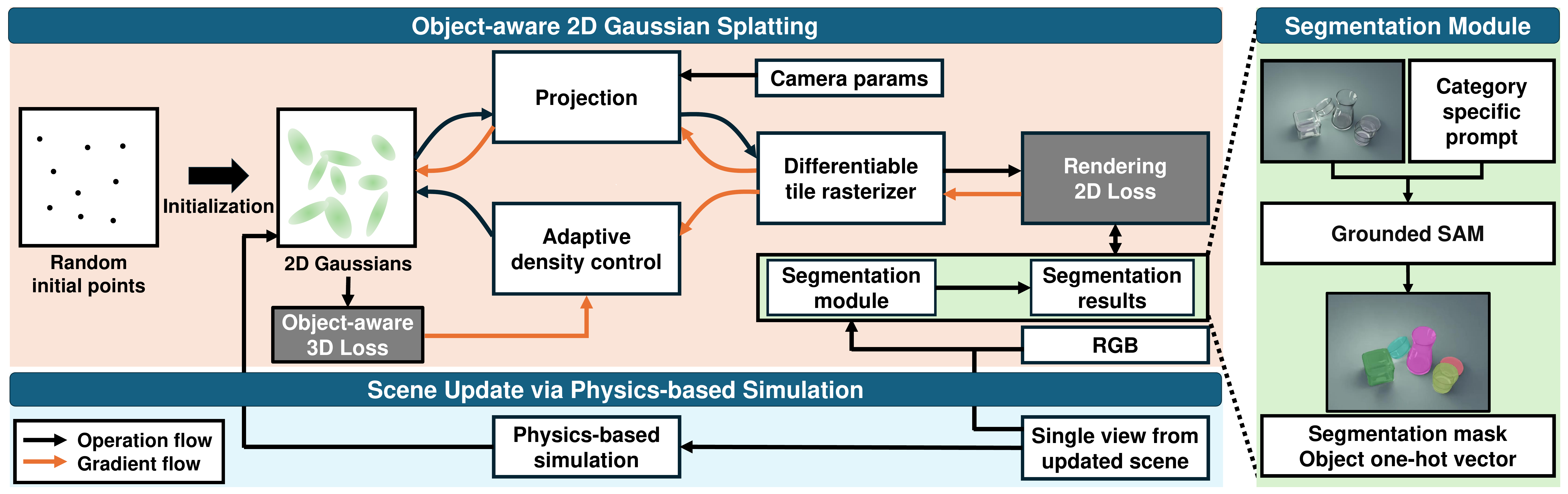}
    \vspace{-3mm}
    \caption{Overview of TRAN-D. First, transparent objects are segmented from sparse views (\textcolor{green}{Section 3.1}). Then, with 2D Gaussians randomly initialized, the process advances through differentiable tile rasterization leveraging segmentation data from the segmentation module and an object‐aware 3D Loss to produce a reliable, fully reconstructed object surface (\textcolor{orange!80!yellow}{Section 3.2}). Finally, the scene is updated via physics-based simulation for object removal and movement (\textcolor{cyan}{Section 3.3}).}
    \label{fig:overview}
    \vspace{-6mm}
\end{figure*}
\section{Related works}
\label{sec:related}

\subsection{Sparse-view Novel View Synthesis for GS}

Sparse-view \ac{NVS} is a critical challenge in 3D reconstruction, aiming to reduce the number of input views from dozens to just a few. 
In the context of \ac{GS}, existing methods address this challenge by distilling additional information into the 2D/3D Gaussians or proposing techniques for efficient optimization. 
Existing methods rely on pre-trained backbones \cite{chen2024mvsplat, wewer2024latentsplat}, leverage 3D foundation models \cite{tang2024mv, fan2024instantsplat}, or use depth priors from monocular depth estimation models \cite{huang2025fatesgs, xiong2023sparsegs, zhu2024fsgs, li2024dngaussian}. However, they
often fail to provide accurate results for transparent objects, and pre-trained models can encounter domain gaps with the available training data, leading to suboptimal performance. In contrast, TRAN-D avoids the reliance on additional networks by introducing object-aware loss, improving performance specifically for transparent object depth reconstruction.

\subsection{Object Reconstruction Using 2D/3D GS}

Recent advancements in \ac{GS} have driven progress in object reconstruction. In this line of study, \ac{3DGS} has been widely employed to represent object geometry, leveraging surface properties (\eg, normals) to model object surfaces \cite{guedon2024sugar, wang2024GausSurf}. However, 3D Gaussians are better suited for volumetric representation, and their multi-view inconsistent nature makes them less effective for accurate surface modeling.

In contrast, \ac{2DGS} \cite{huang20242d} has proven to be better suited for surface modeling, as it directly splats onto the object’s surface, providing more accurate and view-consistent geometry \cite{huang20242d}. By collapsing the 3D volume into 2D oriented planar Gaussian disks, 2D Gaussians offer a more geometrically faithful representation of object surfaces, enhancing the accuracy of the reconstruction. In \cite{rogge2025object}, a method is introduced where segmentation masks are used along with a background loss to better delineate the object. We take this finding further by incorporating object-specific information directly during the optimization process. By splatting segmentation masks and object index one-hot matrices alongside the 2D Gaussians, we not only separate objects from the background but also ensure clear delineation between multiple objects within a scene. 

\subsection{Transparent Object Depth Reconstruction} 

Recent efforts in transparent object depth reconstruction have predominantly followed two streams, NeRF and \ac{GS}. NeRF-based methods \cite{ichnowski-2022-corl, kerr-2022-corl, lee2023nfl} aim to model the scene's radiance field. While being effective, these approaches generally require a large number of training images and suffer from slow training speeds. In particular, Residual-NeRF \cite{duisterhof2024residualnerf} critically depends on the presence of a background image, which can be a significant limitation in many applications. 

\ac{GS}-based methods have also been applied to transparent-object reconstruction. TranSplat \cite{jykim-2025-icra} uses diffusion to generate rich surface features, and TransparentGS \cite{transparentgs} 
 models reflection and refraction via separate BSDFs. While both capture fine surface details, their optimization requires more time, and neither addresses the core limitations of the need for dense multi-view inputs.

\section{Methods}
\label{sec:methods}
As illustrated in \figref{fig:overview}, TRAN-D consists of three modules. 
First, the \textit{\textbf{segmentation module}} leverages Grounded SAM trained with category-specific prompting strategy to isolate transparent object instances. 
Second, the \textit{\textbf{object-aware \ac{2DGS} module}} employs a novel object-aware loss to produce dense and artifact-free reconstructions. 
Finally, the \textit{\textbf{scene update module}} uses physics simulation to predict and refine the reconstruction when objects are removed.



\begin{figure}[t]
  \centering
  \includegraphics[width=0.85\linewidth]{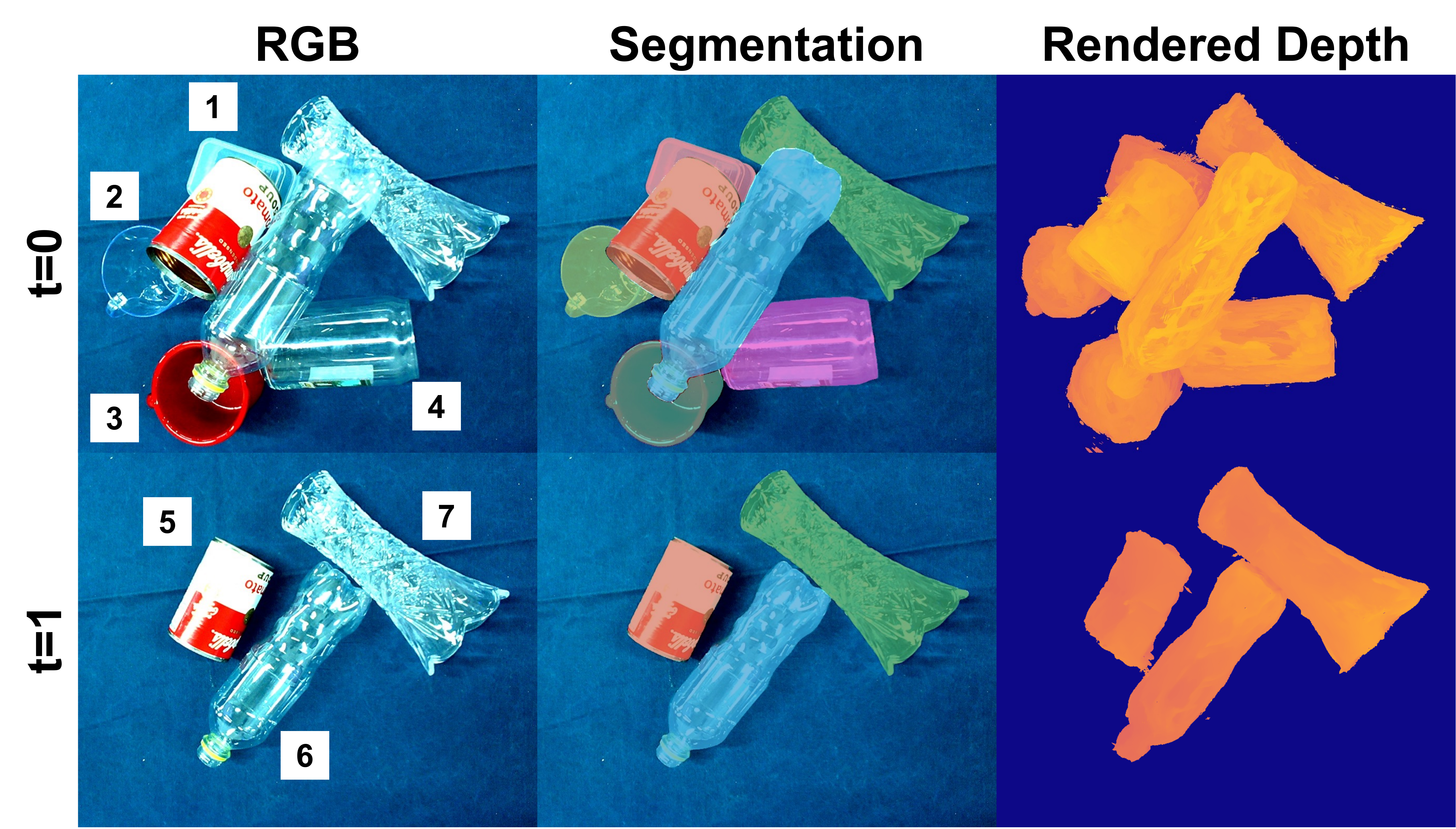}
     \vspace{-2mm}
   \caption{Segmentation and depth rendering result for cluttered scene with both transparent and opaque unseen objects. Upper objects (5 \& 6) topple after removing the lower four.}
   \vspace{-6mm}
   \label{fig:mix}
\end{figure}

\subsection{Transparent Object Segmentation}
\label{sec:method_seg}

Existing segmentation models have difficulty handling cluttered scenes with transparent objects due to occlusions, underscoring the need for specialized training. To overcome this limitation, we fine-tune Grounded SAM \cite{ren2024grounded} by incorporating text prompts alongside image inputs for transparent object segmentation. Inspired by the object-specific prompts used in DreamBooth \cite{ruiz2023dreambooth} and GaussianObject \cite{yang2024gaussianobject}, we integrate similar prompt into training, detailed further in \apref{sec:imp_details}. Since the purpose of segmentation in this work is to assist the \ac{2DGS} in recognizing transparent objects, we do not require distinct object classes. Instead, all transparent objects are treated as a single category and assigned a unique identifier as a category-specific prompt. As a result, it ensures consistent instance segmentation masks across multiple views as shown in \figref{fig:mix} and \apref{sec:seg_more}.


\subsection{Object-aware 2D Gaussian Splatting}
\label{sec:method_2dgs}

In scenes with transparent objects, \ac{SFM} methods \cite{schoenberger2016sfm} often fail to recover reliable points, causing to reconstruction collapse due to poor initialization. This issue also affects 3D foundation models, as seen in InstantSplat \cite{fan2024instantsplat}. 
To overcome this issue, we initialized 2D Gaussians from random points and incorporated additional guidance to enable robust optimization in scenes with transparent objects.
Specifically, we render and compare a combination of RGB images, instance segmentation masks, and object index one-hot vectors in the \ac{2DGS} process.

\begin{figure}[t]
    \centering
    \includegraphics[width=0.8\linewidth]{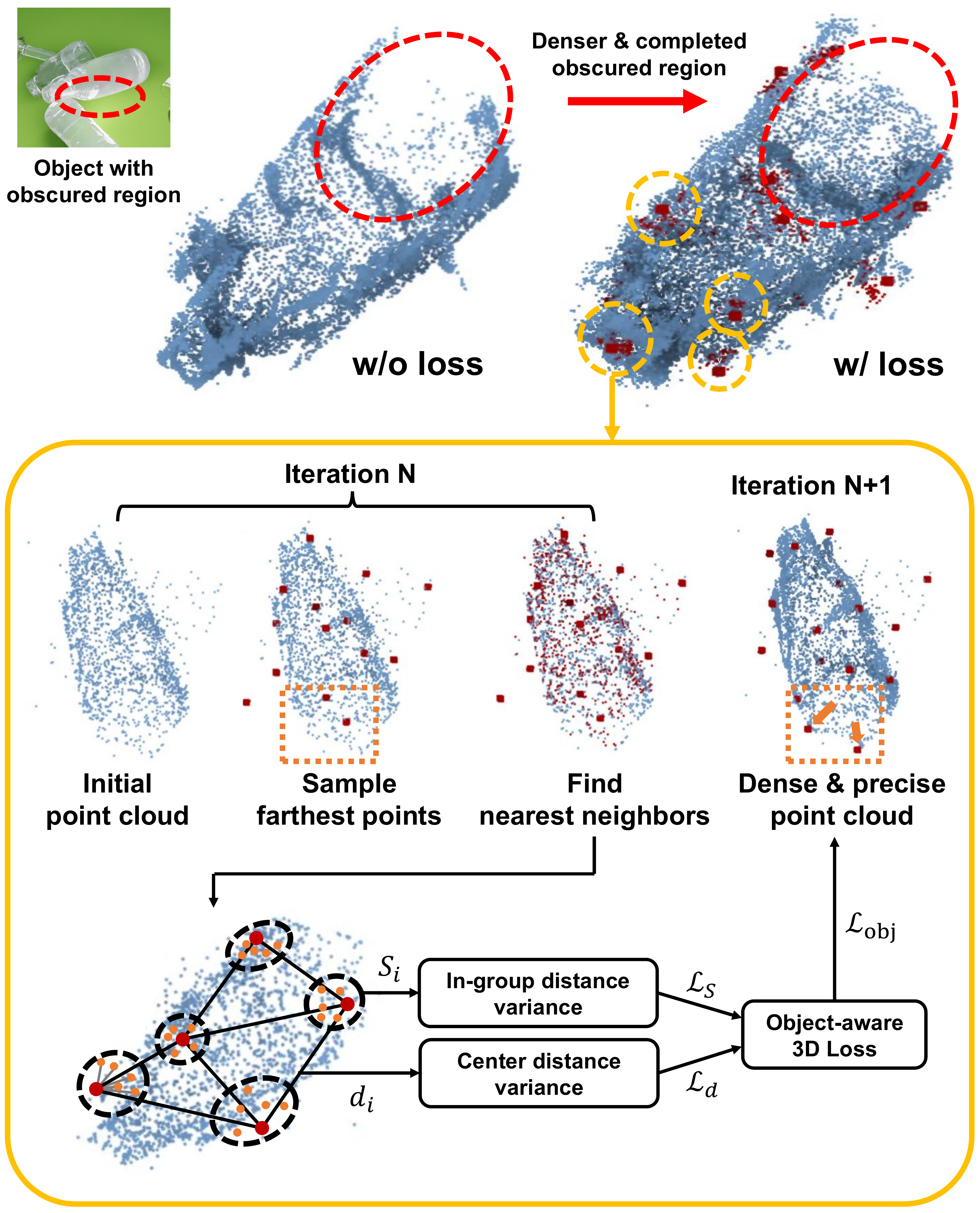}
    \vspace{-2mm}
    \caption{
    Comparison of 2D Gaussian means at without (top left) and with (top right) our object‐aware 3D loss, showing denser coverage in obscured regions. The bottom workflow demonstrates a repeated process of sampling the farthest points, finding their nearest neighbors, and computing a 3D loss. 
    }
    \label{fig:loss}
    \vspace{-6mm}
\end{figure}

In addition, we introduce an object‐aware 3D loss to improve optimization as shown in \figref{fig:loss}. This loss is calculated based on 3D distances both intra-group and inter-group among the 2D Gaussians, effectively regularizing their positions. By employing a hierarchical design that is robust to optimization progress with varying numbers of Gaussians, points can be placed even in fully obscured regions, resulting in a denser and more uniform distribution across the entire object surface.

\subsubsection{Segmentation Mask Rendering}

Let $\mathbf{M} \in \mathbb{R}^{3 \times H \times W}$ be a colorized segmentation mask for a single view as shown in \figref{fig:mix}, where each pixel encodes the segmented object in RGB. Each Gaussian $\mathcal{G}_i$ is assigned a corresponding color vector $\mathbf{m}_i \in \mathbb{R}^3$ representing its associated object. When projecting onto the image plane, the rendered mask $m(x)$ is computed by accumulating each Gaussian's contribution using the modified Gaussian function $\hat{\mathcal{G}}_i(u(x))$ as:
\small
\begin{equation}
m(x) = \sum_{i=1} m_i \alpha_i \hat{\mathcal{G}}_i(u(x)) \prod_{j=1}^{i-1} (1 - \alpha_j \hat{\mathcal{G}}_j(u(x))),
\end{equation}
\normalsize
where $\alpha$ is opacity and $\hat{\mathcal{G}}(u(x))$ is the modified Gaussian function from \ac{2DGS} \cite{rogge2025object}. In addition to color rendering, an object segmentation mask is also rendered, and the Gaussians' object color vectors are optimized with the rendered and ground-truth masks. This prevents the opacity of Gaussians representing transparent objects from collapsing to zero during training, allowing \ac{2DGS} to accurately represent them.

\subsubsection{Object Index One-Hot Vector Rendering}

For scenes with multiple transparent objects, we keep an \emph{object index one-hot vector} $\mathbf{o}_i \in \mathbb{R}^{N+1}$ for each pixel, where $N$ represents the number of objects, and the extra dimension accounts for the background. Analogous to the segmentation mask, we associate each Gaussian $\mathcal{G}_i$ with $\mathbf{o}_i$, indicating the object it belongs to. The rendering equation for the one-hot vector is given by:
\small
\begin{equation}
\hat{\mathbf{o}}(x) \;=\; \sum_{i} \mathbf{o}_i \,\alpha_i \hat{\mathcal{G}}_j(u(x))\prod_{j=1}^{i-1} \left(1 - \alpha_j \hat{\mathcal{G}}_j(u(x))\right).
\end{equation}
\normalsize
The Gaussian-splatted one-hot features $\hat{\mathbf{O}}$ are unbounded by default. To constrain these outputs and ensure valid object index predictions, we apply a softmax activation to each channel for normalization across the object index channels. We then compute a dice loss $\mathcal{L}_{\mathrm{one\text{-}hot}}$ \cite{milletari2016v}  between $\hat{\mathbf{O}}(x)$ and the one-hot labels $\mathbf{O}(x)$ from Grounded SAM.

\subsubsection{Object-aware Loss for Obscured Regions}

In cluttered scenes with limited viewpoints, occlusions often create obscured regions that are not visible from any view, resulting in very weak gradients from rendering. Therefore, relying solely on view-space position gradients can lead to poorly optimized Gaussians.
To address this, we introduce an object-aware loss that generates gradients for obscured Gaussians, guiding the optimization process to make complete surface of the object.

We begin by selecting \( n_g \) the most distant 2D Gaussians for each object—identified via our object index splatting—which serve as center Gaussian of the group. Each group is formed by including its \( n_n \) nearest‐neighbor 2D Gaussians belonging to the same object.
First, we enforce uniform spacing among the group center Gaussian's mean (\( c_i \)) themselves. For each \( c_i \), we compute the minimal distance to all other \( c_j \):
\small
\begin{equation}
d_i = \min_{j \in [1,n_{g}], j \neq i} \| c_j - c_i \|,
\end{equation}
\normalsize
and define the distance variance loss as:
\small
\begin{equation}
\mathcal{L}_{\text{d}} = \operatorname{Var}\left(d_1, d_2, \ldots, d_{n_g}\right).
\end{equation}
\normalsize

This loss helps to anchor Gaussians to the surface of the object, particularly in regions that are obscured. In regions directly visible from the view, 2D Gaussians settle onto the surface like a covering layer, and barely move due to their confident positioning. In contrast, for obscured regions, Gaussians can be located anywhere within the large volume, since the loss does not reflect changes in their positions.
Therefore, $d_i$ values from visible regions remain almost unchanged while others from obscured regions vary considerably. By using their variance as a loss, we encourage the larger, fluctuating distances to approach the stable ones. Ultimately, the centers shift to form appropriately convex surfaces in these obscured regions, a far more reliable and realistic outcome than having them drift too far or become floaters.


Next, for each group \( G_{i} \) \( (1 \le i \le n_g) \), we compute the sum of distance \( S_{i} \) between \( c_i \) and \( n_n \) nearest neighbor Gaussians' mean:
\small
\begin{equation}
S_i = \sum_{x \in \mathrm{NN}(c_i)} \| x - c_i \|.
\end{equation}
\normalsize

To address sparsity in obscured regions we encourage these sums to remain uniform. By promoting consistent local density, TRAN-D can densify the representation in less visible areas of the object. Consequently, previously uncovered areas arising from sparse-view constraints can still attract sufficient Gaussians, ensuring a denser and more robust reconstruction of the entire surface. We formulate this criterion as:

\small
\begin{equation}
\mathcal{L}_{\text{S}} = \operatorname{Var}\left(S_1, S_2, \ldots, S_{n_g}\right).
\end{equation}
\normalsize
To optimize the placement of Gaussians effectively across the entire process, we implement a three‐level hierarchical grouping strategy. In the beginning, only a small number of Gaussians exist for each object because the optimization begins from random points and simultaneously learns the object’s one‐hot index. At this early phase, using too many groups can cause overlapping neighborhoods that reduce efficiency. Later, as more Gaussians appear for each object, having too few groups diminishes the advantage of grouping. Therefore, we employ three different (\( n_g \), \( n_n \)) configurations, ensuring that the loss function remains both meaningful and effective throughout all stages of optimization.

The overall object‐aware 3D loss is obtained by aggregating the losses from each object at each hierarchical level:
\small
\begin{equation}
\mathcal{L}_{\text{obj}} = \sum_{l=1}^{3}\sum_{o=1}^{N}(a_{S}\mathcal{L}_{S} + a_{d}\mathcal{L}_{d}).
\end{equation}
\normalsize
The final optimize loss function is given by:
\small
\begin{align}
\mathcal{L} =\; & a_{\text{color}}\,\mathcal{L}_{\mathrm{c}} + a_{\text{mask}}\,\mathcal{L}_{\mathrm{m}} + a_{\text{one-hot}}\,\mathcal{L}_{\mathrm{one\text{-}hot}} + \mathcal{L}_{\mathrm{obj}},
\end{align}
\normalsize
where \(\mathcal{L}_{\mathrm{c}}\) is the RGB reconstruction loss, combining L1 loss with the D-SSIM term as in \cite{kerbl20233d}. Similarly \(\mathcal{L}_{\mathrm{m}}\) is formulated in a similar manner for segmentation mask, combining L1 loss with the D-SSIM term. We set the following hyperparameters:
$a_{\text{color}} = 0.5$, $a_{\text{mask}}=0.5$,  $a_{\text{one-hot}} = 1.0$, $a_{S} = 10000/3$, $a_{d} = 1/3$.  For each level in the hierarchical grouping, we assign the pairs (16,16), (32,16), (64,32) as the (\( n_g \), \( n_n \)) values.


\begin{figure}[t]
    \centering
    \includegraphics[width=0.9\linewidth]{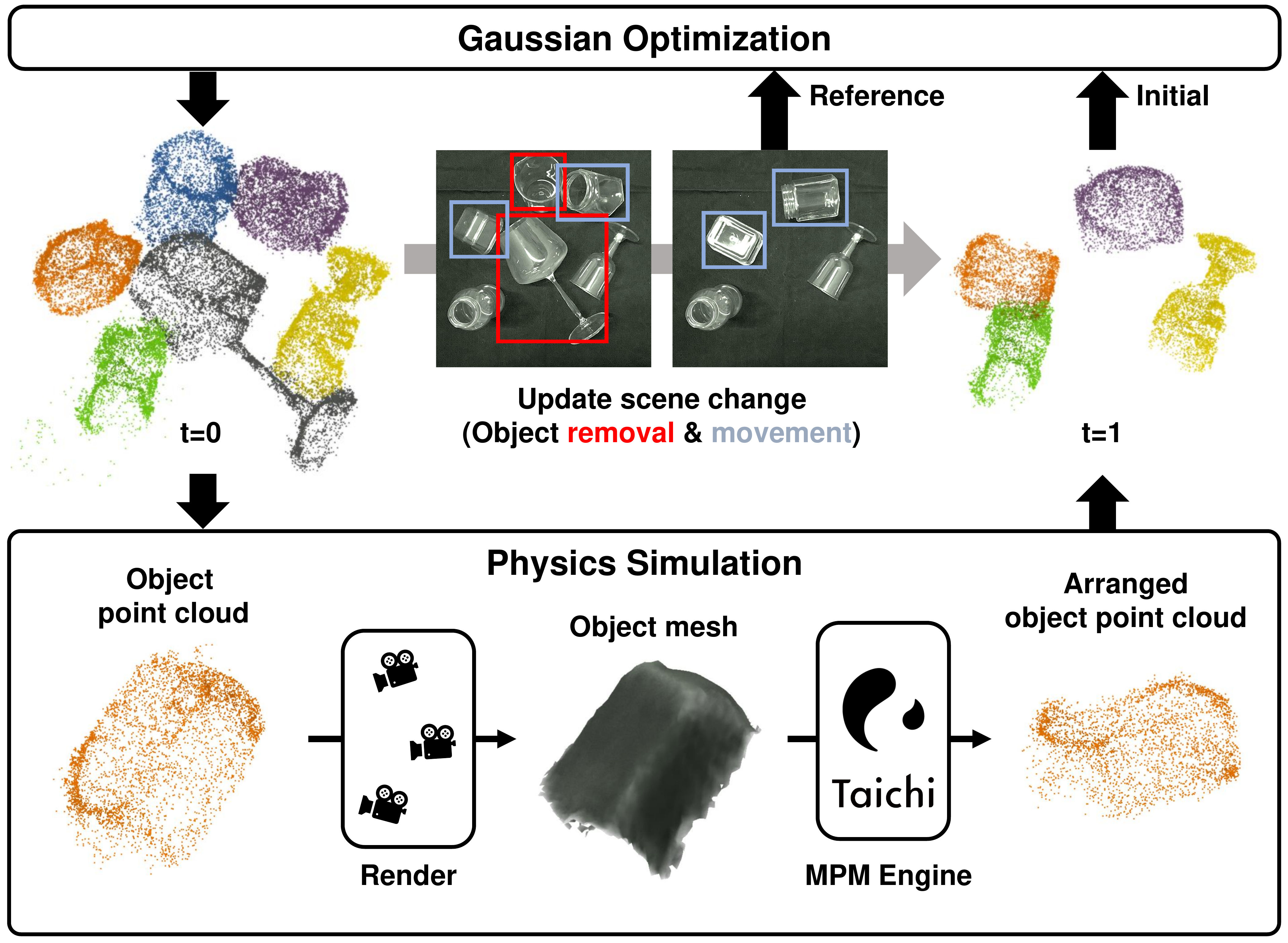}
    \vspace{-2mm}
    \caption{Overview of the scene update process using physics simulation. Starting with object Gaussians, a corresponding object mesh is generated by rendered depth. 
    The MPM engine exploits mesh to simulate positional shift of objects, updating the scene from $t=0$ to $t=1$. Finally, a single image is used for Gaussian optimization, ensuring scene changes are accurately reflected in the 2D Gaussian representation.}
    \label{fig:simulation}
    \vspace{-6mm}
\end{figure}

\subsection{Scene update via Physic-based Simulation}

Since the proposed method has strong surface reconstruction capability that enables robust physics simulations, we can reliably update scene dynamics, as shown in \figref{fig:simulation}.

When an object is removed from the scene, we first perform segmentation using fine-tuned Grounded SAM to identify the object from the previous state. The corresponding Gaussians are isolated using the object index one-hot vector and subsequently removed.

Next, we render a depth map from the prior 2D Gaussian representation to generate a mesh. This mesh is essential for the physics simulation because it provides the necessary surface points for accurately modeling dynamics.

The scene is then updated by simulating the effects of object removal using the material-point method (MPM) implemented in Taichi \cite{hu2019difftaichi}. This simulation captures the chain-reaction movement among multiple neighboring objects, ensuring precise scene updates. Because the physics simulation does not directly yield a perfect Gaussian representation, we re-optimize the Gaussian splatting process to refine the scene. Notably, during this re-optimization, we omit the object-aware loss since it was already applied in the initial optimization to ensure that the object surfaces were accurately represented. Further details can be found in \apref{sec:imp_details}.

\section{Experiments}
\label{sec:experiments}
\begin{table*}[ht]
    \centering
    \caption{Depth reconstruction results for synthetic TRansPose. For each $t$, best results highlighted in \textbf{bold}; Second best in \underline{underlines}.}
    \vspace{-2mm}
    \label{tab:real-transpose}
    \resizebox{0.85\textwidth}{!}{%

\begin{tabular}{c|c|cccccccc}
\hline
             & \textbf{}                  & \textbf{MAE $\downarrow$}    & \textbf{RMSE $\downarrow$}   & \textbf{$\delta$ \textless 0.5cm $\uparrow$} & \textbf{$\delta$ \textless 1cm $\uparrow$} & \textbf{$\delta$ \textless 2.5cm $\uparrow$} & \textbf{$\delta$ \textless 5cm $\uparrow$} & \textbf{$\delta$ \textless 10cm $\uparrow$} & \textbf{$\delta$ \textless 20cm $\uparrow$} \\ \hline
 & 3DGS             & 0.0965          & 0.1161          & 4.22\%                         & 8.49\%                       & 20.68\%                        & 35.36\%                      & 57.17\%                       & 89.68\%                       \\
             & 2DGS             & 0.0691    & \textbf{0.0914} & 6.14\%                   & 12.96\%                & 32.27\%                  & 51.24\%                & 74.23\%                 & {\ul 95.12\%}                 \\
             & InstantSplat      & 0.1605          & 0.1900          & 1.61\%                         & 3.35\%                       & 10.22\%                        & 28.02\%                      & 52.62\%                       & 68.07\%                       \\
\textbf{t=0}             & FSGS              & 0.1702          & 0.2079          & 1.08\%                         & 2.13\%                       & 5.42\%                         & 10.95\%                      & 27.89\%                       & 70.90\%                       \\
             & Feature Splatting & 0.0915          & 0.1287          & 5.10\%                         & 10.20\%                      & 25.59\%                        & 44.56\%                      & 68.11\%                       & 90.68\%                       \\ 
             & TranSplat         & {\ul 0.0632}          & {\ul 0.0982}          & {\ul 8.14\%}                         & {\ul 16.92\%}                       & {\ul 43.01\%}                        & {\ul 62.85\%}                      & {\ul 77.42\%}                       & 93.91\%                       \\
             & NFL         & 0.1932          & 0.2269          & 1.80\%                         & 3.64\%                       & 9.59\%                        & 19.63\%                      & 31.85\%                       & 50.58\%                       \\
             & Dex-NeRF         & 0.4096          & 0.4260          & 0.13\%                         & 0.25\%                       & 0.65\%                        & 1.35\%                      & 2.63\%                       & 5.10\%                       \\
            \cline{2-10} 
             & \textbf{Ours}              & \textbf{0.0380} & 0.1069    & \textbf{13.40\%}               & \textbf{29.30\%}             & \textbf{69.11\%}               & \textbf{89.15\%}             & \textbf{95.96\%}              & \textbf{97.37\%}              \\ \hline \hline
 & 3DGS             & 0.1132          & 0.1311    & 4.11\%                         & 8.26\%                       & 19.82\%                        & 30.55\%                      & 47.94\%                       & 83.34\%                       \\
             & 2DGS             & \textbf{0.0849} & \textbf{0.1083} & 5.14\%                   & 10.25\%                & 25.05\%                  & 42.10\%                & 65.82\%                 & \textbf{91.60\%}              \\
             & InstantSplat      & 0.1688          & 0.1904          & 2.61\%                         & 5.12\%                       & 13.66\%                        & 32.13\%                      & 52.98\%                       & 64.78\%                       \\
\textbf{t=1}             & FSGS              & 0.1422          & 0.1672          & 1.89\%                         & 3.79\%                       & 9.46\%                         & 18.08\%                      & 38.95\%                       & 75.69\%                       \\
             & Feature Splatting & 0.1556          & 0.1988          & 3.27\%                         & 6.56\%                       & 16.27\%                        & 28.86\%                      & 46.64\%                       & 68.36\%                       \\ 
             & TranSplat         & 0.0879          & {\ul 0.1169}          & {\ul 6.44\%}                         & {\ul 13.11\%}                       & {\ul 31.62\%}                        & {\ul 49.19\%}                      & {\ul 67.01\%}                       & 86.46\%                       \\
             & NFL         & 0.2047          & 0.2356          & 1.85\%                         & 3.74\%                       & 9.57\%                        & 18.53\%                      & 34.10\%                       & 61.91\%                       \\
             & Dex-NeRF         & 0.4120          & 0.4283          & 0.11\%                         & 0.22\%                       & 0.56\%                        & 1.12\%                      & 2.40\%                       & 5.43\%                       \\
             \cline{2-10} 
             & \textbf{Ours}              & {\ul 0.0864}    & 0.1971          & \textbf{8.39\%}                & \textbf{17.50\%}             & \textbf{48.46\%}               & \textbf{77.08\%}             & \textbf{88.70\%}              & {\ul 90.76\%}                \\\hline
\end{tabular}
    }

\end{table*}

\begin{table*}[ht]
    \centering
    \caption{Depth reconstruction results for synthetic ClearPose. For each $t$, best results highlighted in \textbf{bold}; Second best in \underline{underlines}.
    \vspace{-2mm}
    }
    \label{tab:real-clearpose}
    \resizebox{0.85\textwidth}{!}{%

\begin{tabular}{c|c|cccccccc}
\hline
             & \textbf{}                  & \textbf{MAE $\downarrow$}    & \textbf{RMSE $\downarrow$}   & \textbf{$\delta$ \textless 0.5cm $\uparrow$} & \textbf{$\delta$ \textless 1cm $\uparrow$} & \textbf{$\delta$ \textless 2.5cm $\uparrow$} & \textbf{$\delta$ \textless 5cm $\uparrow$} & \textbf{$\delta$ \textless 10cm $\uparrow$} & \textbf{$\delta$ \textless 20cm $\uparrow$} \\ \hline
 & 3DGS             & 0.1358          & 0.1703          & 3.94\%                         & 7.84\%                       & 18.37\%                        & 30.60\%                      & 47.63\%                       & 73.56\%                       \\
             & 2DGS             & 0.1091          & 0.1452          & 4.91\%                         & 9.86\%                       & 24.12\%                        & 41.09\%                      & 62.20\%                       & 81.55\%                       \\
             & InstantSplat               & 0.1764          & 0.2143          & 2.37\%                         & 4.81\%                       & 12.39\%                        & 25.83\%                      & 44.66\%                       & 65.63\%                       \\
\textbf{t=0}             & FSGS              & 0.1562          & 0.1768          & 0.77\%                         & 1.57\%                       & 4.46\%                         & 10.84\%                      & 28.83\%                       & 71.56\%                       \\

             & Feature Splatting & {\ul 0.0801}    & \textbf{0.1046} & 5.19\%                         & 10.53\%                      & 25.51\%                        & 44.71\%                      & {\ul 69.28\%}                 & {\ul 92.54\%}\\
                          & TranSplat         & 0.0905          & 0.1280          & {\ul 6.62\%}                   & {\ul 13.58\%}                & {\ul 31.95\%}                  & {\ul 51.18\%}                & 68.53\%                       & 84.89\%                       \\
                          & NFL         & 0.1441          & 0.1847          & 2.65\%                   & 5.30\%                & 13.34\%                  & 26.32\%                & 45.22\%                       & 68.24\%                       \\
                          & Dex-NeRF         & 0.3933          & 0.4161          & 0.26\%                   & 0.53\%                & 1.32\%                  & 2.64\%                & 5.30\%                       & 11.74\%                       \\
              \cline{2-10} 
             & \textbf{Ours}              & \textbf{0.0461} & {\ul 0.1047}    & \textbf{10.54\%}               & \textbf{22.42\%}             & \textbf{54.38\%}               & \textbf{76.53\%}             & \textbf{93.18\%}              & \textbf{97.67\%}              \\ \hline \hline
 & 3DGS             & 0.1571          & 0.1890          & 2.92\%                         & 5.73\%                       & 12.78\%                        & 21.51\%                      & 37.56\%                       & 67.75\%                       \\
             & 2DGS             & 0.1263          & 0.1637          & {\ul 4.99\%}                   & {\ul 9.87\%}                 & {\ul 22.19\%}                  & 35.44\%                      & 55.19\%                       & 77.47\%                       \\
             & InstantSplat               & 0.1850          & 0.2230          & 2.52\%                         & 5.06\%                       & 12.56\%                        & 25.23\%                      & 42.16\%                       & 60.07\%                       \\
\textbf{t=1}             & FSGS              & 0.1452          & 0.1723          & 1.95\%                         & 3.88\%                       & 9.69\%                         & 19.83\%                      & 38.43\%                       & 73.08\%                       \\
             & Feature Splatting & {\ul 0.0995} & \textbf{0.1266} & 4.68\%                         & 9.28\%                       & 21.67\%                        & {\ul 37.17\%}                & {\ul 59.53\%}                 & {\ul 86.56\%}\\  
             & TranSplat         & 0.1221          & {\ul 0.1560}    & 4.73\%                         & 9.50\%                       & 22.00\%                        & 36.28\%                      & 53.34\%                       & 77.42\%\\
             & NFL         & 0.1410          & 0.1790          & 2.55\%                  & 5.16\%                & 13.54\%                  & 27.68\%                & 47.21\%                       & 70.33\%                       \\
                          & Dex-NeRF         & 0.4060          & 0.4310          & 0.25\%                   & 0.49\%                & 1.24\%                  & 2.46\%                & 4.83\%                       & 10.20\%                       \\
              \cline{2-10} 
             & \textbf{Ours}              & \textbf{0.0910}    & 0.1899         & \textbf{6.87\%}                & \textbf{14.07\%}             & \textbf{36.47\%}               & \textbf{64.37\%}             & \textbf{84.02\%}              & \textbf{91.41\%}             \\\hline
\end{tabular}%
    }
\vspace{-2mm}
\end{table*}

\subsection{Experimental Setup}
\noindent \textbf{Dataset} We conducted experiments in both synthetic and real-world environments. Since no existing benchmark dataset includes transparent object removal sequences, we created synthetic sequences with various backgrounds and textures for quantitative evaluation. Using models from the transparent object datasets, we generated 9 sequences of unseen objects with ClearPose \cite{chen-2022-eccv} and 10 sequences with TRansPose \cite{kim2024transpose} using BlenderProc \cite{Denninger2023}. To construct a realistic scene after removing objects, we applied physics-based simulation using BlenderProc’s built-in physics engine to refine the object poses. 

For synthetic data, we captured 6 images at 60-degree intervals along the Z-axis, and all models used these images for optimization. For other baselines, 6 images from the post state were used for training, while our approach utilized single bird-eye view image. For each state, we captured images from 30 random poses and used them as test images.

For the real-world experiments, We captured 6 real world sequences including both seen and unseen, transparent and opaque objects using a Franka Emika Panda arm and a RealSense L515, recording RGB images and ground-truth poses. Although quantitative evaluation was not possible, all baselines used nine views for both the initial and post-change states, whereas TRAN-D required only a single bird’s-eye view for post-change refinement.

\noindent \textbf{Metric} We evaluate the performance of TRAN-D using three primary metrics: depth accuracy, training time, and the number of Gaussians. For depth accuracy, we compare the rendered depth against the ground truth object depth using several evaluation metrics, including Mean Absolute Error (MAE), Root Mean Squared Error (RMSE), and threshold percentages at various depth thresholds ( $<$ 0.5 cm, $<$1 cm, $<$2.5 cm, $<$5 cm, $<$10 cm, $<$20 cm). All comparisons are performed using absolute depth values, allowing for a direct comparison of depth accuracy across methods. 
To gauge TRAN-D’s efficiency, we compare both the total training duration—including preprocessing and optimization—and the number of Gaussians used to represent the scene. Details of the implementation for segmentation, Gaussian optimization, and physics simulation are provided in \apref{sec:imp_details}.
\subsection{Baselines}
We compare TRAN-D with existing approaches that target scene reconstruction. These include 3D Gaussian Splatting  \cite{kerbl20233d} and 2D Gaussian Splatting \cite{huang20242d}, which are effective for scene reconstruction but face challenges in sparse-view settings. Additionally, we compare with Feature Splatting \cite{qiu-2024-featuresplatting}, which utilizes foundation models like CLIP \cite{radford2021learning}, DINO \cite{oquab2023dinov2}, and SAM \cite{kirillov2023segment} for feature extraction. We also look at methods such as InstantSplat \cite{fan2024instantsplat}  and FSGS \cite{zhu2024fsgs}, which rely on foundation models for sparse-view optimization. Finally, we compare TRAN-D with TransSplat \cite{jykim-2025-icra}, Dex-NeRF \cite{ichnowski-2022-corl} and NFL \cite{lee2023nfl}, specifically designed for transparent object reconstruction.

For the object removal scenario, we use the Gaussian means from $t=0$ as the initial points at $t=1$ to update the scene. However, InstantSplat does not perform densification and emphasizes rapid scene reconstruction, so even at $t=1$, it reinitializes Gaussians with 3D foundation model. Additionally, we provided the ground‐truth pose and disabled the pose optimization for InstantSplat.

\begin{figure*}[t]
    \centering
    \includegraphics[width=0.93\linewidth]{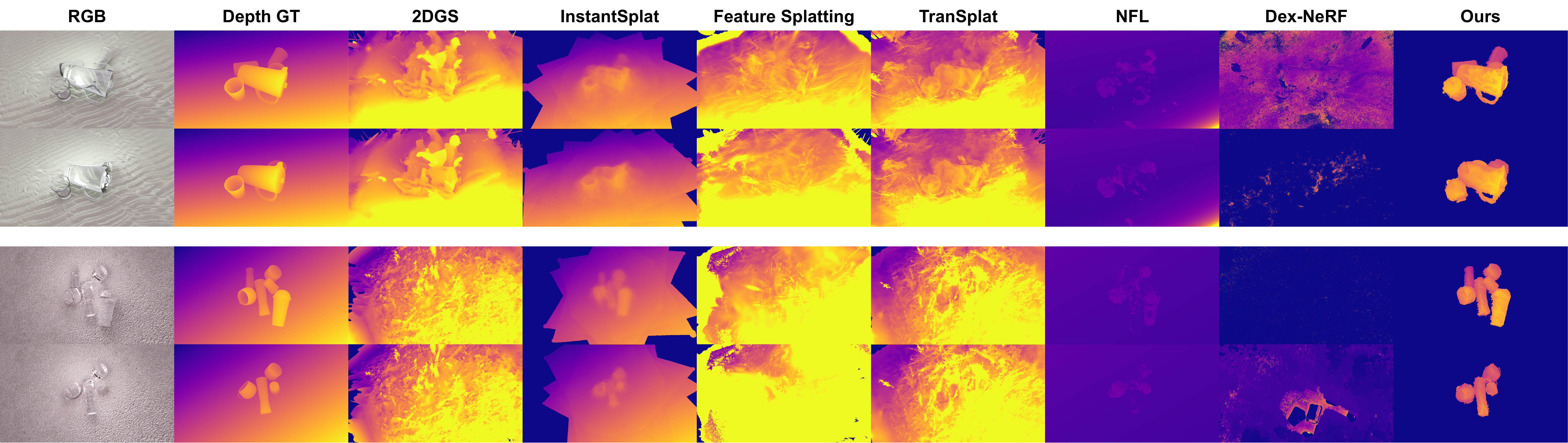}
    \vspace{-2mm}
    \caption{Depth reconstruction results of synthetic sequences. First row is $t=0$, second row is $t=1$.}
    \label{fig:syn}
    \vspace{-4mm}
\end{figure*}
\begin{figure*}[t]
    \centering
    \includegraphics[width=0.93\linewidth]{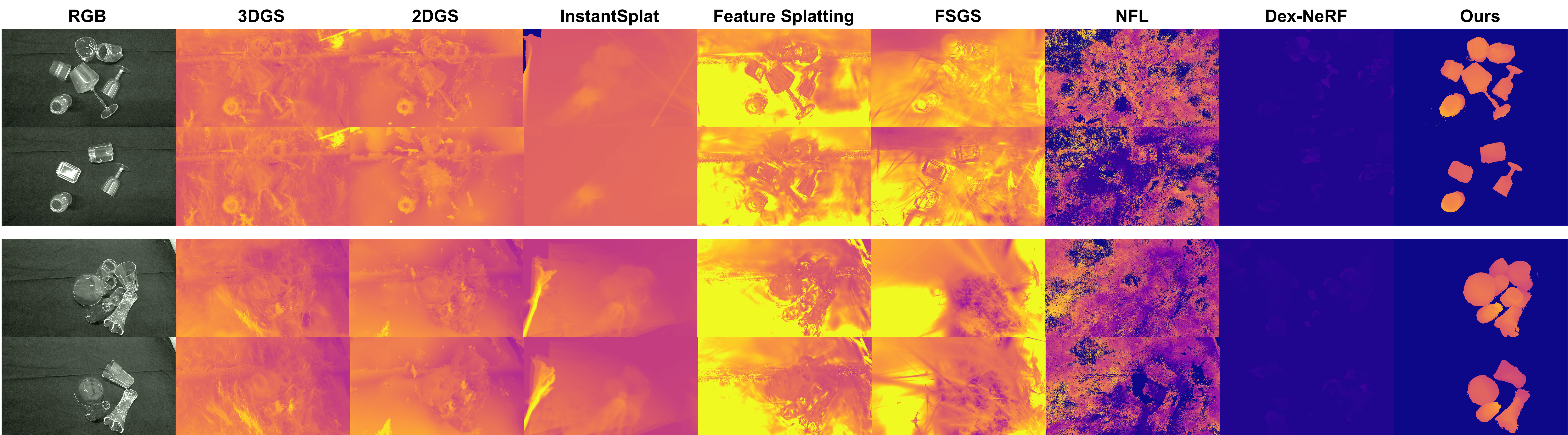}
    \vspace{-2mm}
    \caption{Depth reconstruction results of real-world sequences. First row is $t=0$, second row is $t=1$.}
    \label{fig:real}
    \vspace{-4mm}
\end{figure*}
\subsection{Depth Reconstruction}

Unlike other models that include the entire scene during rendering, TRAN-D renders only the objects. As shown in \tabref{tab:real-transpose} and \tabref{tab:real-clearpose}, TRAN-D achieves the best depth reconstruction performance, outperforming all baselines in terms of MAE and threshold percentage on the TRansPose and ClearPose synthetic sequences at $t=0$. This improvement can be attributed to our approach, which removes the background and focuses on optimizing the object's Gaussians using segmentation masks and object index splatting, resulting in enhanced depth accuracy. Even when only one image is available for refinement at $t=1$, TRAN-D maintains excellent performance, further highlighting the impact of physics-based simulation in refining depth accuracy.
\begin{table}[t]
    \centering
    \caption{Efficiency comparison of baseline methods, using average results from 19 scenes in ClearPose and TRansPose. We evaluate training time and the number of Gaussians, incorporating each method’s specific preprocessing—InstantSplat’s 3D foundation model initialization and feature extraction steps in TranSplat and Feature Splatting. At $t=1$, our method’s time includes physics simulation.}
    \vspace{-2mm}
    \label{tab:ablation_time}     
    \resizebox{0.9\columnwidth}{!}{

\begin{tabular}{cc|ccc|c}
\hline
\multicolumn{1}{l}{}              & \multicolumn{1}{l|}{}     & \multicolumn{3}{c|}{Training time}                                                    & \multicolumn{1}{c}{Gaussians} \\ \cline{3-5}
                                  &                           & Preprocess $\downarrow$ & Perform $\downarrow$ & \multicolumn{1}{c|}{Total $\downarrow$} & \multicolumn{1}{c}{count $\downarrow$}                                 \\ \hline
\multicolumn{1}{c|}{}             & 3DGS            & -                 & 344.1            & 344.1                               & 175.2k                                               \\
\multicolumn{1}{c|}{}             & 2DGS            & -                 & 440.9            & 440.9                               & 227.8k                                               \\
\multicolumn{1}{c|}{}             & InstantSplat     & 22.8                & 56.0             & 78.8                                & 850.1k                                               \\
\multicolumn{1}{c|}{t=0} & FSGS             & -                 & 1476.1           & 1476.1                              & 57.8k                                                \\
\multicolumn{1}{c|}{}             & TranSplat        & 126.6               & 469.5            & 596.0                               & 297.8k                                               \\
\multicolumn{1}{c|}{}             & Feature Splatting & 5.9                & 294.2            & 334.6                               & 88.5k                                                \\ \cline{2-6} 
\multicolumn{1}{c|}{}             & \textbf{Ours}             & \textbf{5.2}        & \textbf{48.9}    & \textbf{54.1}                       & \textbf{33.5k}                                                \\ \hline \hline
\multicolumn{1}{c|}{}             & 3DGS            & -                 & 401.3            & 401.3                               & 266.1k                                               \\
\multicolumn{1}{c|}{}             & 2DGS            & -                 & 447.6            & 447.6                               & 248k                                                 \\
\multicolumn{1}{c|}{}             & InstantSplat     & 28.6                & 66.9             & 95.5                                & 987.2k                                               \\
\multicolumn{1}{c|}{t=1} & FSGS             & -                 & 417.0            & 417.0                               & 52.3k                                                \\
\multicolumn{1}{c|}{}             & TranSplat        & 101.0                & 511.7            & 612.7                               & 318.4k                                               \\
\multicolumn{1}{c|}{}             & Feature Splatting & 5.8                & 221.3            & 259.8                               & 84.5k                                                \\ \cline{2-6} 
\multicolumn{1}{c|}{}             & \textbf{Ours}             & \textbf{10.5}                & \textbf{3.3}              & \textbf{13.8}                                & \textbf{16k}                                                  \\ \hline
\end{tabular}
}
\vspace{-4mm}
\end{table}
In contrast, models like Feature Splatting, InstantSplat, and FSGS, which rely on foundation models, often struggle with transparent objects. These models fail to distinguish transparent objects from the background, leading to artifacts in the rendered output and overall poor performance. Similarly, TranSplat, which uses diffusion-based depth reconstruction, also fails to remove artifacts and performs poorly in sparse-view conditions.

From a qualitative results, as shown in \figref{fig:syn} and \apref{sec:app_more}, Feature Splatting, 2DGS, and TranSplat produce many artifacts. InstantSplat likewise faces challenges, producing depth estimates that nearly coincide with the floor level. In the real‐world sequences, as shown in Fig. \ref{fig:real}, these problems persist. Compared to other models, TRAN-D can capture even thin object parts—such as a cup’s handle—demonstrating its ability to recover fine details and deliver accurate depth reconstruction in complex scenes.

\subsection{Efficiency}
As shown in \tabref{tab:ablation_time}, TranSplat suffers from long preprocessing times due to the computational complexity of diffusion model. 
Similarly, 3DGS, 2DGS, and FSGS also demonstrate the common issue of extended training times inherent in Gaussian Splatting. InstantSplat achieves faster training than other baselines, but its reliance on a 3D foundation model yields an excessively large number of initial points, leading to an overabundance of Gaussians.

In contrast, TRAN-D offers a distinct advantage in terms of efficiency. By separating objects from the background, the number of Gaussians used is significantly smaller compared to these baseline methods. Additionally, the object‐aware loss prevents the formation of floaters and keeps the Gaussian count minimal, preserving accurate depth reconstruction and supporting faster optimization. At $t=0$, TRAN-D achieves results in under one minute, and at $t=1$, the scene update requires only 13.8 seconds. The reduction in Gaussian count also leads to a decrease in optimization time. This demonstrates the efficiency of TRAN-D in both training time and computational cost.

\subsection{Ablation study}

\subsubsection{Analysis on Sparse View}
\begin{table}[h]
\centering

\caption{Ablation study on numbers of views}    
\vspace{-2mm}
    \label{tab:ablation_3}  

\resizebox{\columnwidth}{!}{
\begin{tabular}{cc|cc|cc|cc}
\hline
                               & \textbf{}     & \multicolumn{2}{c|}{3-Views}                           & \multicolumn{2}{c|}{6-Views}                                         & \multicolumn{2}{c}{12-Views}                           \\ \cline{3-8} 
\textbf{}                      & \textbf{}     & \multicolumn{1}{c|}{MAE $\downarrow$}             & RMSE $\downarrow$           & \multicolumn{1}{c|}{MAE $\downarrow$}                           & RMSE $\downarrow$            & \multicolumn{1}{c|}{MAE $\downarrow$}             & RMSE $\downarrow$            \\ \hline
\multicolumn{1}{c|}{}          & InstantSplat  & \multicolumn{1}{c|}{0.1306}          & 0.1727          & \multicolumn{1}{c|}{0.1682}                        & 0.2020          & \multicolumn{1}{c|}{0.2062}          & 0.2343          \\ \cline{2-8} 
\multicolumn{1}{c|}{t=0}       & FSGS          & \multicolumn{1}{c|}{0.1846}          & 0.2147          & \multicolumn{1}{c|}{0.1636}                        & 0.1931          & \multicolumn{1}{c|}{0.1426}          & 0.1792 \\ \cline{2-8} 
\multicolumn{1}{c|}{\textbf{}} & \textbf{Ours} & \multicolumn{1}{c|}{\textbf{0.0405}} & \textbf{0.0968} & \multicolumn{1}{c|}{\textbf{0.0419}}               & \textbf{0.1059} & \multicolumn{1}{c|}{\textbf{0.0448}} & \textbf{0.1154} \\ \hline \hline
\multicolumn{1}{c|}{}          & InstantSplat  & \multicolumn{1}{c|}{0.1539}          & 0.1880          & \multicolumn{1}{c|}{0.1630}                        & 0.1959          & \multicolumn{1}{c|}{0.2033}          & 0.2283          \\ \cline{2-8} 
\multicolumn{1}{c|}{t=1}       & FSGS          & \multicolumn{1}{c|}{0.1570}          & 0.1862          & \multicolumn{1}{c|}{0.1436} & 0.1696          & \multicolumn{1}{c|}{0.1074}          & \textbf{0.1466} \\ \cline{2-8}
\multicolumn{1}{c|}{\textbf{}} & \textbf{Ours} & \multicolumn{1}{c|}{\textbf{0.0706}} & \textbf{0.1621} & \multicolumn{1}{c|}{\textbf{0.0926}}               & \textbf{0.1637} & \multicolumn{1}{c|}{\textbf{0.0953}} & 0.2053 \\ \hline
\end{tabular}
}
\vspace{-4mm}
\end{table}

To evaluate TRAN-D’s robustness to varying numbers of training images, we conducted experiments on the synthetic dataset using 3, 6, and 12 training views. We compared TRAN-D against InstantSplat and FSGS, which also target sparse-view reconstruction. \tabref{tab:ablation_3} shows that the depth accuracy of TRAN-D remains relatively stable, even as the number of training views changes. Qualitative results can be found in \apref{sec:app_viewnum}.

\subsubsection{Object-aware Loss and Physics Simulation}
\begin{table}[ht!]
\centering

\caption{Ablation study on loss}  
\vspace{-2mm}
    \label{tab:ablation_2}  

\resizebox{0.8\columnwidth}{!}{
\begin{tabular}{cc|cc|c}
\hline
\multicolumn{1}{c}{}                    &               & \multicolumn{2}{c|}{Depth accuracy}         & Gaussians      \\ \cline{3-4}
\multicolumn{1}{c}{}                   &               & MAE $\downarrow$            & \multicolumn{1}{c|}{RMSE $\downarrow$} & counts $\downarrow$        \\ \hline
\multicolumn{1}{c|}{\multirow{2}{*}{t=0}}  & w/o object-aware loss      & 0.0447          & 0.1136                    & 35983          \\
\multicolumn{1}{c|}{}                    & \textbf{Full model} & \textbf{0.0419} & \textbf{0.1059}           & \textbf{33482} \\ \hline\hline
\multicolumn{1}{c|}{}                    & w/o object-aware loss      & 0.0932          & 0.2011                    & 16835          \\
\multicolumn{1}{c|}{t=1}                  & w/o simulation       & 0.0891          & 0.1945                    & 15976          \\
\multicolumn{1}{c|}{}                     & \textbf{Full model} & \textbf{0.0886} & \textbf{0.1936}           & \textbf{15974}  \\ \hline
\end{tabular}%
}
\vspace{-4mm}
\end{table}


We conducted an ablation study to evaluate the individual contributions of our object-aware loss and physics simulation. The object-aware loss is designed to guide Gaussians toward obscured regions of the object, improving overall coverage. As shown in \tabref{tab:ablation_2}, including the object-aware loss reduces both MAE and RMSE, and further decreases the number of Gaussians used, indicating that the model reconstructs more of the object’s surface with fewer but better-optimized Gaussians.

\begin{figure}[t!]
    \centering
    \includegraphics[width=0.85\linewidth]{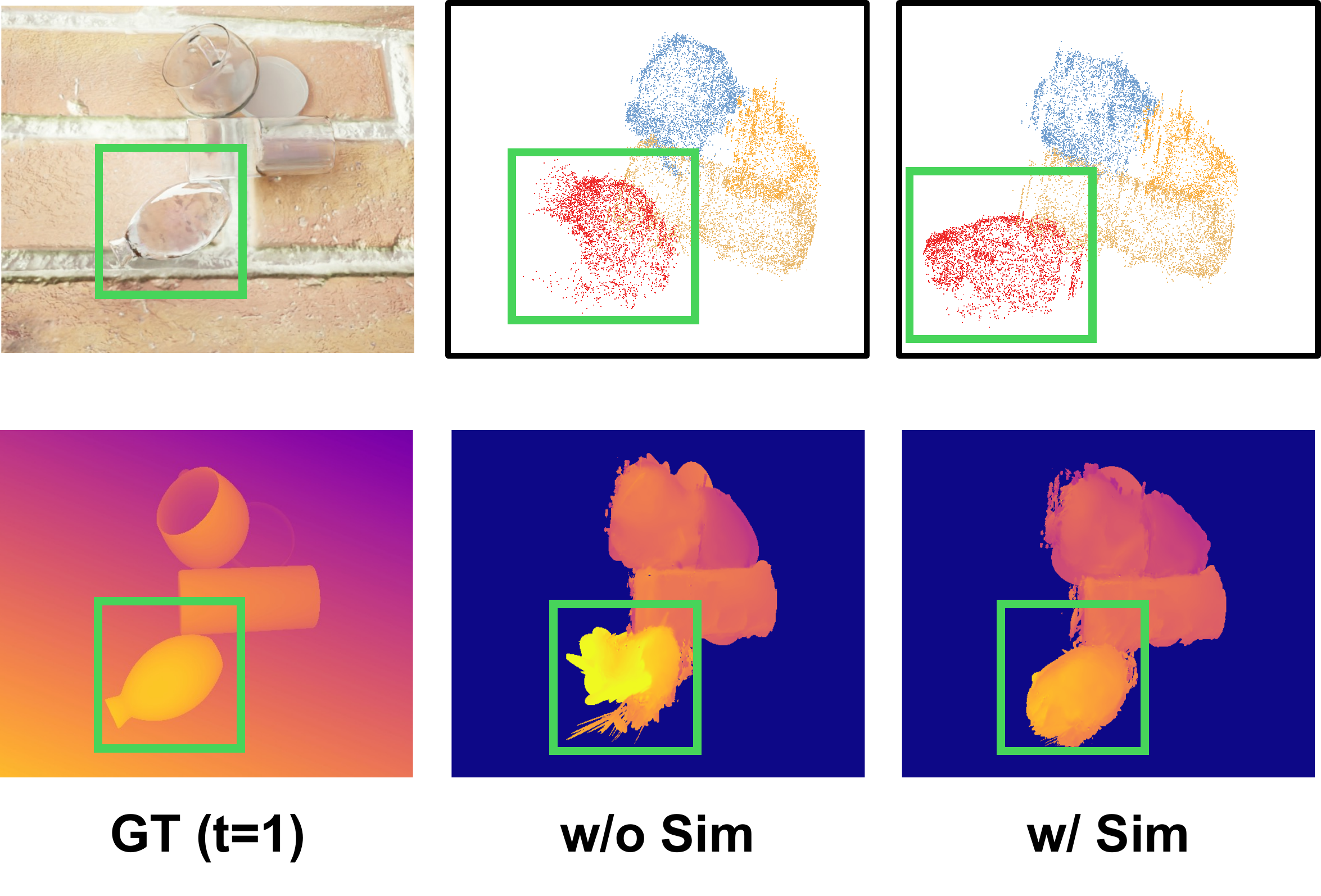}
    \vspace{-4mm}
    \caption{Depth rendering results after object removal and re-optimization. The object within the green box moves in the post-removal state. Without simulation(center), the position of the Gaussian does not move along the Z-axis, leading to failure in accurate depth reconstruction. In contrast, With simulation(right), the Gaussian position is adjusted, resulting in a more accurate and consistent depth representation.
    }
    \label{fig:sim_no_sim}
    \vspace{-6mm}
\end{figure}

The physics simulation influences the reconstruction at $t=1$, when transitioning from $t=0$. We observe that incorporating physics simulation further reduces both MAE and RMSE, demonstrating its effectiveness in updating the scene. As shown in \figref{fig:sim_no_sim}, omitting physics simulation often leads to overfitting to the training images at $t=1$, causing the object to lose its shape. By contrast, physics simulation preserves object geometry, emphasizing its crucial role.

\vspace{-1mm}

\section{Conclusion}
\label{sec:conclusion}
\vspace{-1mm}
Although dense depth reconstruction for transparent objects has been actively studied through neural rendering techniques, existing methods often require substantial training time, dense-view inputs, and do not account for object dynamics. In this paper, we presented TRAN-D, a physics simulation-aided sparse-view 2D Gaussian Splatting approach combined with transparent object segmentation masks, enabling accurate depth reconstruction within a minute. Moreover, we introduced an object-aware loss that influences obscured regions, thereby improving depth accuracy while also reducing training time and the total number of Gaussians required compared to previous methods. 

Despite these advantages, TRAN-D remains heavily dependent on segmentation quality. As shown in \apref{sec:app_fail}, tracking failures, intense lighting, or backgrounds that make object boundaries difficult to delineate can degrade performance. Additionally, TRAN-D currently handles only partial object removal or slight movements. Future work will focus on addressing these limitations by developing a more robust, segmentation-independent approach capable of handling more complex dynamics and lighting environments—extending our method's applicability to a wider range of real-world scenarios.

\section*{Acknowledgement}
This work was supported by the National Research Foundation of Korea (NRF) grant funded by the Korea government (MSIT)(No. RS-2024-00461409), and in part by Hyundai Motor Company and Kia. and the Institute of Information \& communications Technology Planning \& Evaluation (IITP) grant funded by the Korea government(MSIT) No.2022-0-00480, Development of Training and Inference Methods for Goal-Oriented Artificial Intelligence Agents.

{
    \small
    \bibliographystyle{ieeenat_fullname}
    \bibliography{main, rpm_packages/string-short}
}

\appendix
\clearpage
\setcounter{page}{1}
\twocolumn[
  \begin{center}
    {\Large \bfseries TRAN-D: 2D Gaussian Splatting-based Sparse-view Transparent Object \\ Depth Reconstruction via Physics Simulation for Scene Update \par}
    \vspace{0.25cm}
    {\large\bfseries Supplementary Material}
    \vspace{0.5cm}
  \end{center}
]
\section{Implementation Details}
\label{sec:imp_details}
\noindent \textbf{Segmentation} 
Generic prompts like “glass” or “transparent” often misclassify background regions or merge overlapping objects. To avoid these failures, we create an intentionally non-dictionary, category-specific text prompt “786dvpteg” to unambiguously denote “transparent object,” while all other objects use the generic prompt “object.” We did not include the “object” prompt during training. We fine-tuned Grounded DINO for 1 epoch with a batch size of 12, conducted on a single NVIDIA A6000 GPU. The training was performed on a synthetic TRansPose \cite{kim2024transpose} dataset generated using BlenderProc \cite{Denninger2023}. Only the image backbone portion was fine‐tuned, while the text backbone BERT \cite{devlin2019bert} layers remained frozen.

\noindent \textbf{Gaussian Optimization} The Gaussian Splatting model optimization was performed on a single NVIDIA RTX 2080 Ti GPU. The model parameters are mostly identical to those used in the \ac{2DGS} method. We began the optimization process with random points. The learning rate for the object index one-hot vector was initialized at 0.1 and decayed to 0.0025 over 1000 iterations. Following object removal, the scene was refined over the course of 100 iterations.

\noindent \textbf{Physics Simulation} In our physics‐based scene update, we employ the Material Point Method (MPM), which represents bodies as material points carrying mass, momentum, and material properties, and simulates their interactions under gravity and collisions with other objects to compute forces and update positions. 

To generate a suitable particle distribution, we first convert our optimized 2D Gaussian splatting surface into a mesh via surface reconstruction of the rendered depth map. We then sample particles at approximately uniform spacing over this mesh. Each particle is assigned material properties—Young’s modulus \(5 \times 10^{4}\ \mathrm{Pa}\) and Poisson’s ratio 0.4—values selected after evaluating several combinations because this pair consistently yielded the fastest simulation times without any observable loss in dynamic fidelity or visual quality. Other parameter settings provided comparable physical accuracy but incurred higher computational cost. 

We run the MPM simulation for 100 timesteps. At each step, particle masses and velocities are used to compute stresses and body forces including gravity. Collisions are resolved against static geometry, with the ground height set to 0 to account for contact and collisions with the floor, as well as neighboring particles. Particle positions are updated accordingly. After simulation, we perform a brief 100 iterations of Gaussian re‐optimization, omitting the object‐aware 3D loss since surface consistency was already enforced during the initial optimization.
\section{Segmentation Results} 
\label{sec:seg_more}
\begin{figure}[t]
  \centering
  \includegraphics[width=\linewidth]{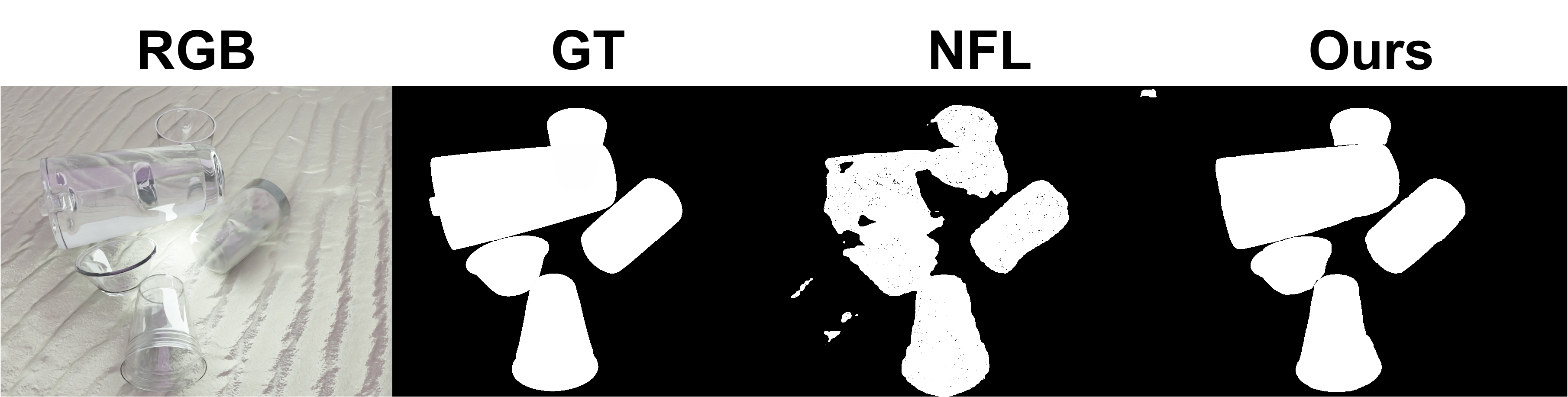}
   \caption{Even in boundary-ambiguous regions, our method delivers accurate object segmentation, whereas the mask-based NFL produces noisy or incomplete masks.}
   \vspace{-4mm}
   \label{fig:seg_compare}
\end{figure}

We compare the segmentation performance of NFL for transparent-object reconstruction (\figref{fig:seg_compare}). Despite targeting transparent objects, these methods struggle when boundaries become ambiguous due to lighting variations, background clutter, or occlusions.

\begin{figure*}[!ht]
  \centering
  \begin{subfigure}{0.49\textwidth}
    \centering
    \includegraphics[width=\textwidth]{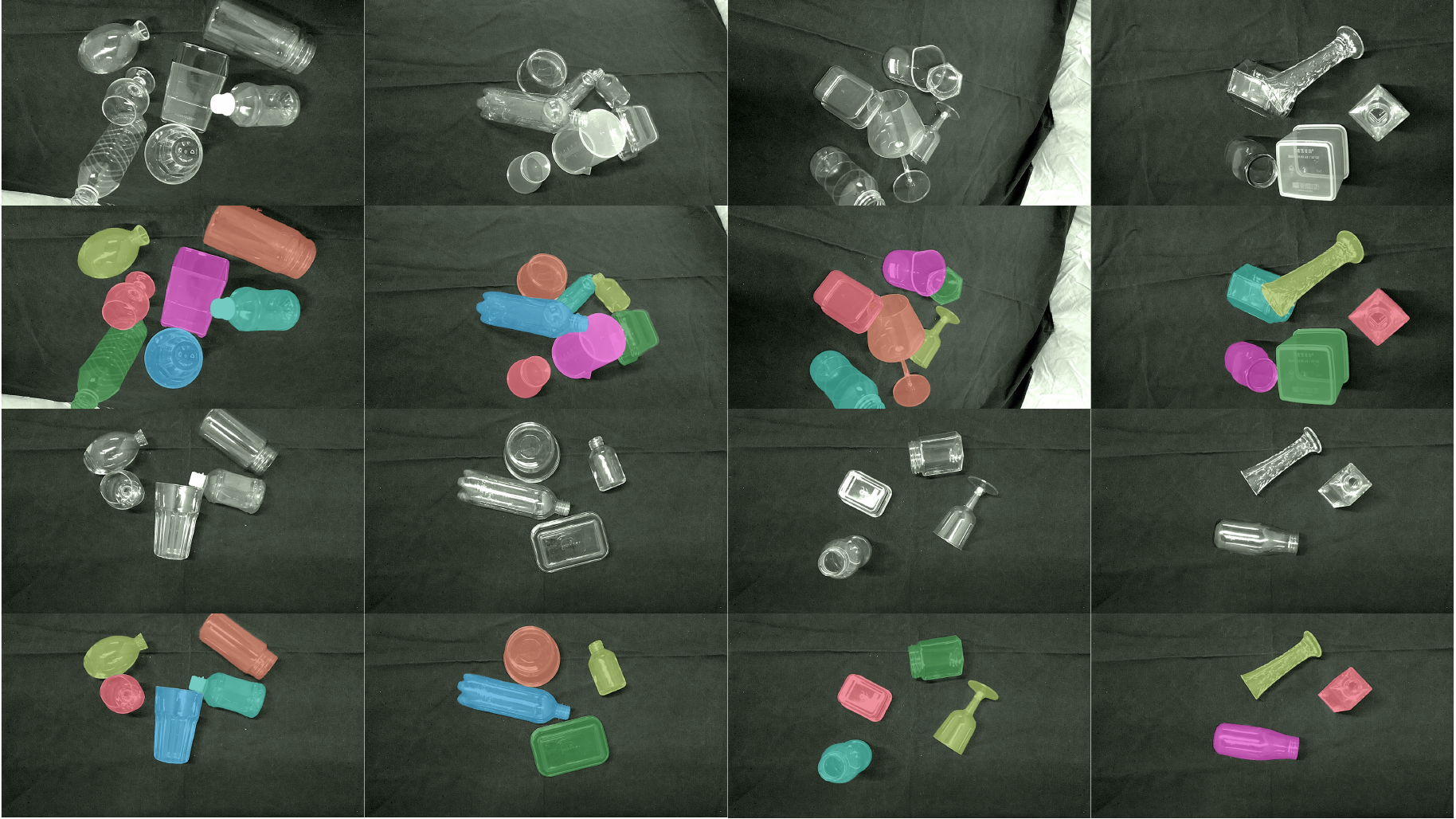}
    \caption{Real-world transparent objects}
    \label{fig:more_real}
  \end{subfigure}
  \begin{subfigure}{0.49\textwidth}
    \centering
\includegraphics[width=\textwidth]{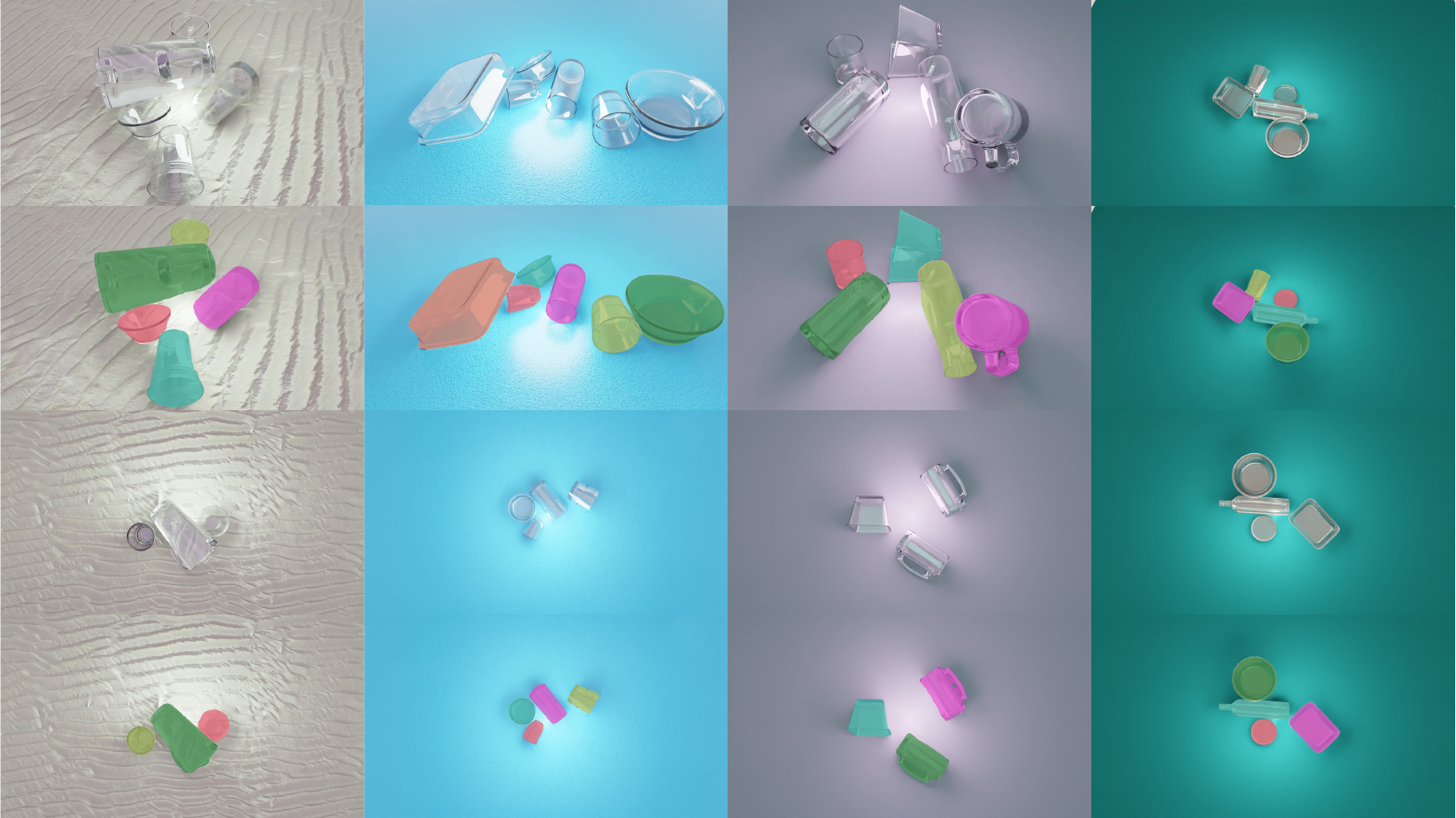}
    \caption{Synthetic transparent objects}
    \label{fig:more_syn}
  \end{subfigure}
  \caption{Segmentation result visualization for transparent objects.
  }
  \label{fig:more_segmenation}
\end{figure*}

Moreover, despite ambiguous boundaries, cluttered backgrounds, and varied viewpoints, our model consistently produces accurate masks for transparent objects across all scenarios (\figref{fig:more_segmenation}).

\section{Additional Qualitative Results}
\label{sec:app_more}

In this supplementary material, we present additional qualitative results highlighting the robustness of TRAN-D.

\begin{figure*}[ht]
    \centering
    \includegraphics[width=0.95\linewidth]{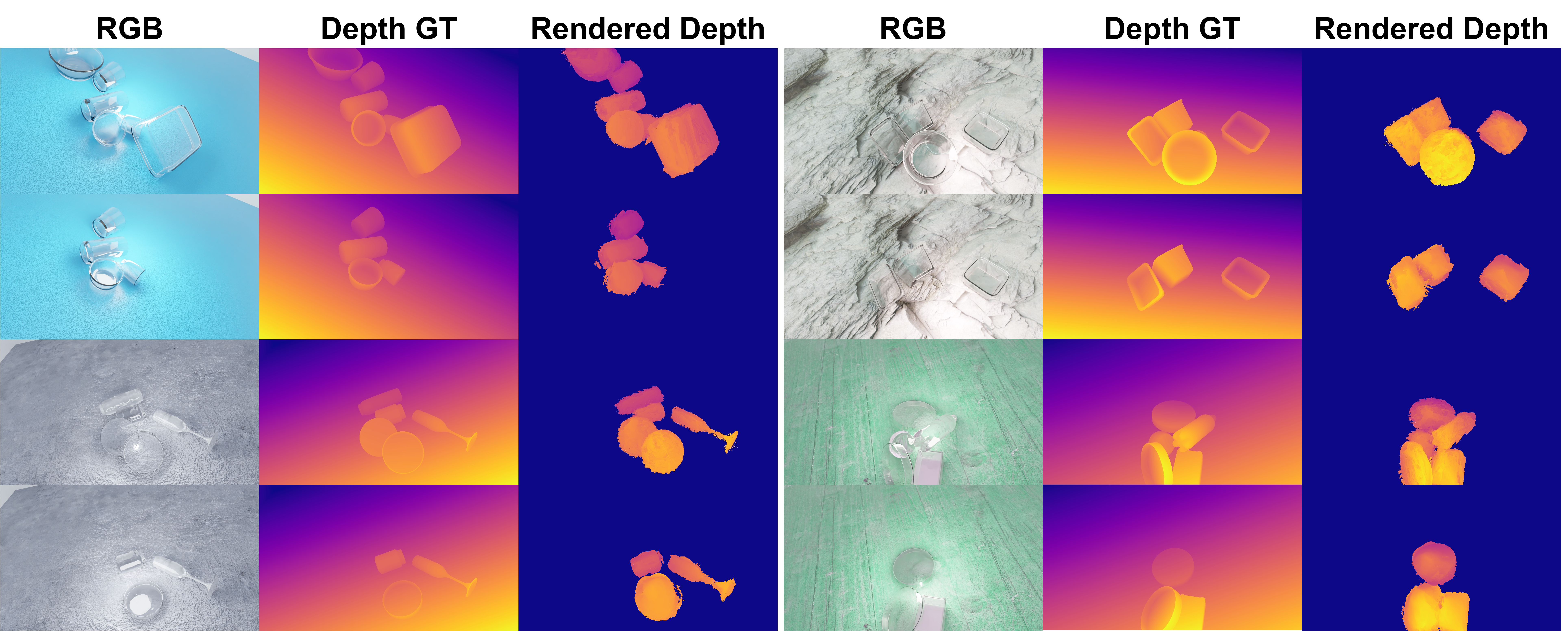}
    \caption{Depth reconstruction results of synthetic sequences.}
    \label{fig:more1}
    \vspace{-2mm}
\end{figure*}

As illustrated in \figref{fig:more1}, TRAN-D consistently produces accurate and stable depth reconstructions across a variety of object types and background textures. Notably, TRAN-D maintains high performance even in challenging scenarios, such as scenes featuring complex, marble-like backgrounds where transparent objects are visually difficult to distinguish, and cases with significant object overlap or occlusion.

\begin{figure*}[ht]
    \centering
    \includegraphics[width=\linewidth]{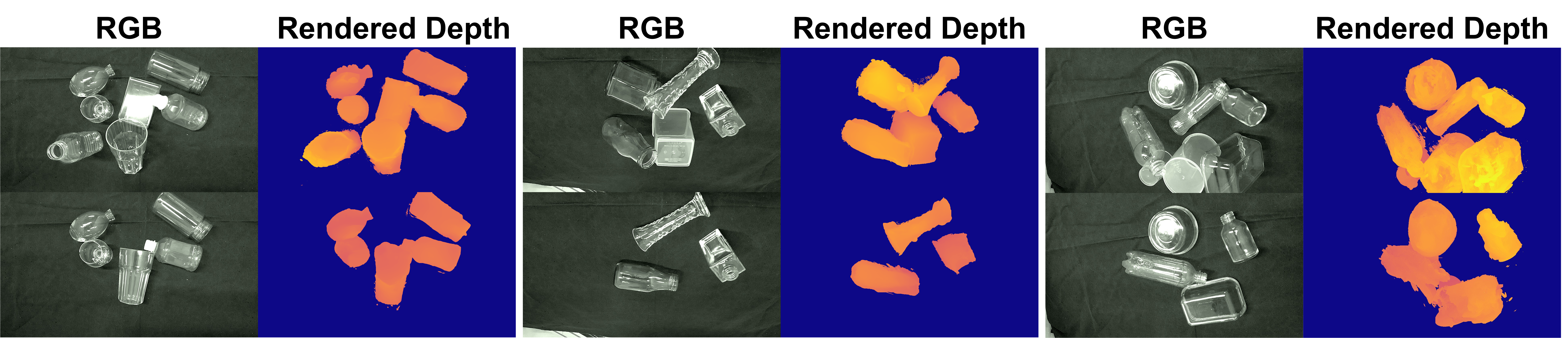}
    \caption{Depth reconstruction results of real-world sequences.}
    \label{fig:more2}
    \vspace{-2mm}
\end{figure*}

Moreover, \figref{fig:more2} presents additional real-world results demonstrating the practicality and effectiveness of TRAN-D in real-world applications. Unlike synthetic scenarios, real-world conditions introduce complexities such as closely placed or interdependent objects, significantly complicating the reconstruction task. Despite these challenges, TRAN-D achieves reliable depth reconstructions, highlighting its potential for use in complex, real-world environments. For comprehensive visualization of results across more scenes, please refer to the supplementary video.

\begin{figure}[ht]
    \centering
    \includegraphics[width=0.95\linewidth]{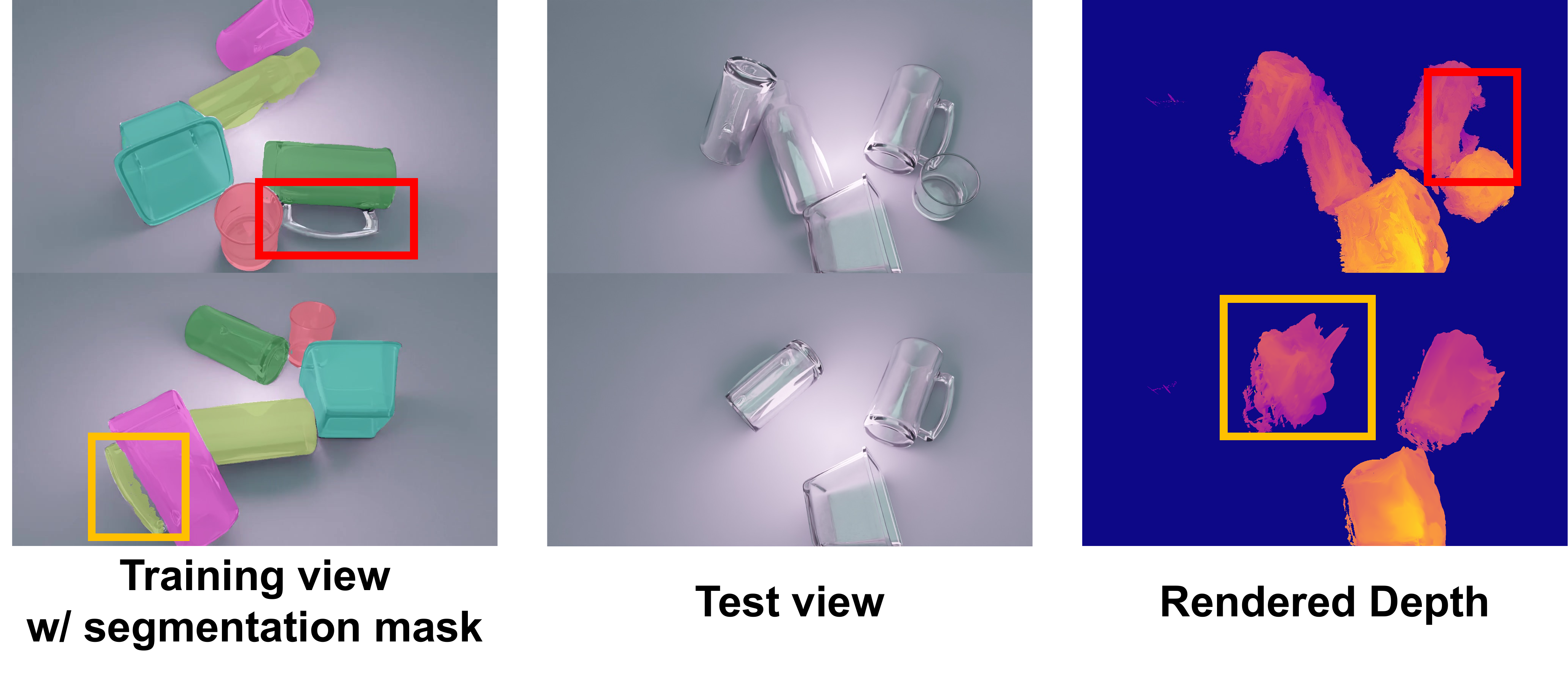}
    \caption{Failure cases due to segmentation inaccuracies. In the first case (yellow box), incorrect labeling in the segmentation process results in poor reconstruction of the pink object, which causes the physics simulation to fail at $t=1$. In the second case (red box), part of the object is not segmented, leading to incomplete reconstruction at $t=0$.}
    \label{fig:failure}
    \vspace{-2mm}
\end{figure}

\section{Qualitative Analysis on the Number of Views}
\label{sec:app_viewnum}

\begin{figure*}[ht]
    \centering
    \includegraphics[width=\linewidth]{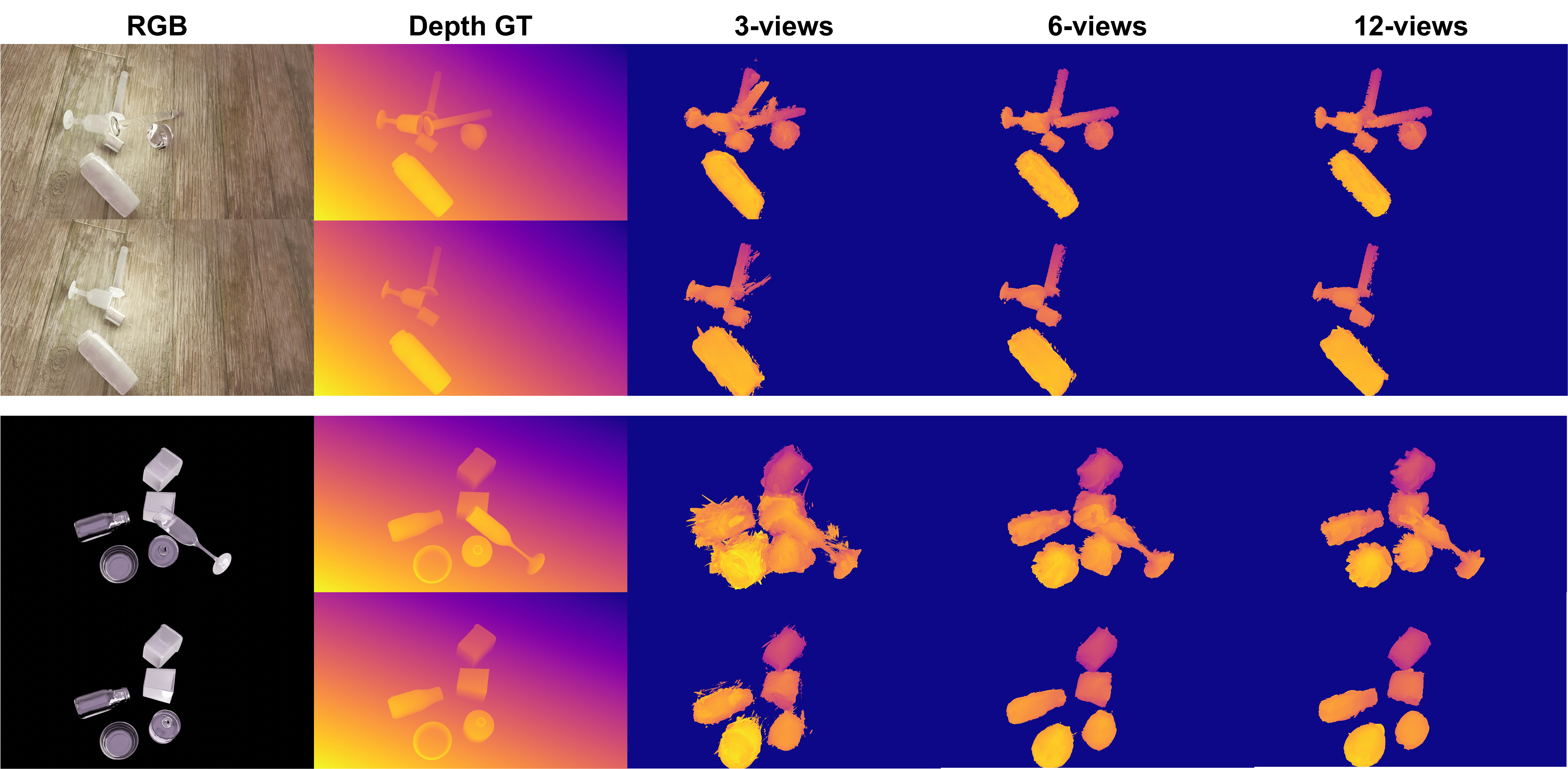}
    \caption{Depth reconstruction results of 3, 6, 12 views in our model.}
    \label{fig:nview}
    \vspace{-2mm}
\end{figure*}

In this section, we analyze the qualitative performance of TRAN-D with different numbers of training views. \figref{fig:nview} illustrates that increasing the number of training views generally leads to better depth reconstruction, with only a minimal difference observed between 6 and 12 views. This indicates that while additional views offer extra information, the model's performance stabilizes after a certain number of views (around 6 in our case).

TRAN-D performs well even with just 6 views, offering reliable depth reconstructions with minimal artifacts. However, increasing the number of views beyond 6 does not result in a substantial improvement, which implies diminishing returns in terms of reconstruction accuracy as the number of views increases.

\section{Failure Cases and Discussion}
\label{sec:app_fail}

Our Grounded SAM, trained on transparent objects, provides object-level segmentation masks that distinguish objects from the background and enable object-aware loss computation. This offers a significant advantage to our model; however, it is also highly sensitive to segmentation quality. 

In sparse-view scenarios, large viewpoint shifts complicate consistent object tracking, and physical properties of transparent objects, such as reflection and refraction under intense lighting, introduce ambiguity in object boundaries. These factors, combined with inadequate segmentation coverage in occluded areas, further degrade reconstruction quality. As shown in \figref{fig:failure}, when parts of an object are not grouped together or are incorrectly segmented as different objects, the reconstruction suffers and the physics simulation may fail.


\end{document}